\documentclass[nonatbib, final,12pt,authoryear]{elsarticle}

\usepackage{amssymb}
\usepackage{placeins}
\usepackage{subcaption}
\usepackage{amsmath}
\usepackage{hyperref}
\usepackage{array}
\usepackage{geometry}

\usepackage{graphicx}
\usepackage{tikz}
\usetikzlibrary{arrows.meta, shapes.geometric, positioning}
\usetikzlibrary{calc} 

\usepackage{hyperref}
\newcommand{\commentblock}[1]{}

\usepackage{enumitem}

\newenvironment{compactitem}{
    \begin{itemize}[leftmargin=*, itemsep=0.5em, parsep=0pt]
}{
    \end{itemize}
}
\newcommand\tabref[1]{Table~\ref{#1}}

\makeatletter
\let\c@author\relax
\makeatother

\makeatletter
\def\ps@pprintTitle{%
 \let\@oddfoot\@empty
 \let\@evenfoot\@empty
}
\makeatother

\usepackage[backend=biber,style=vancouver]{biblatex}
\usepackage[symbol]{footmisc}

\bibliography{references}

\begin{document}

\begin{frontmatter}


\title{Road Graph Generator: Mapping roads at construction sites from GPS data}



\author[1]{Katarzyna Michałowska\footnote{Corresponding author: katarzyna.michalowska@sintef.no}}
\author[1]{Helga Margrete Bodahl Holmestad}
\author[1]{Signe Riemer-Sørensen}

\address[1]{Mathematics and Cybernetics, SINTEF Digital, Oslo, Norway}



\begin{abstract}

We propose a new method for inferring roads from GPS trajectories to map construction sites. This task presents a unique challenge due to the erratic and non-standard movement patterns of construction machinery, which significantly diverge from typical vehicular traffic on established roads. Our proposed method first identifies intersections in the road network that serve as critical decision points, and then connects them with edges to produce a graph, which can subsequently be used for planning and task allocation. The method includes several physically meaningful parameters that can be tuned to adapt it to different scenarios. We demonstrate the approach by mapping roads at a real-life construction site in Norway. The method is validated on four increasingly complex segments of the map. In our tests, the method achieved perfect precision and recall in detecting intersections and inferring roads in data with no or low noise, while its performance was reduced in areas with significant noise and consistently missing GPS updates.
\end{abstract}




\end{frontmatter}


\section{Motivation}
Reducing carbon emissions is a critical goal in mitigating the far-reaching impacts of climate change. 
In Norway, the building and construction sector contributes directly and indirectly to 15\% of total greenhouse gas emissions (2019), with construction vehicles accounting for 1.5\% of the total emissions (2021) \cite{BAE, Veikart}. 

In managing construction projects, it is crucial to effectively handle the complex interplay of resources, including personnel, materials, and machinery, to ensure their optimal interaction in terms of type, quantity, location, and time \cite{Dardouri2023}. 
One viable strategy to address this issue is to maximize the use of heavy machinery, such as dump trucks and excavators, through optimal coordination and task allocation. According to Skanska Norge AS, the largest construction contractor in Norway, mass transporting units are idling 40--60\% of the time during projects. This is a consequence of the conventional planning approach where excavators are considered more expensive than mass transportation units, and hence the latter are excessively deployed to avoid idling excavators, often leading to overcapacity and idling of transportation units instead.  
Improved coordination on construction sites, e.g., optimized real-time task allocation for dumpers, can reduce this overcapacity, leading to lower greenhouse gas emissions and saving both time and money. This is a critical improvement given the increasing size of construction projects involving hundreds of mass transporting units.


The first step in optimizing these operations is to acquire an up-to-date map of the construction site. It is imperative that the map generation be automated given the constantly evolving nature of the construction site and because manual mapping would be impractical and deprioritized, as emphasized in interviews with drivers, foremen, and construction site coordinators. Herein, we address this need by proposing an algorithm for road inference from Global Positioning System (GPS) trajectories recorded by vehicles on the construction site. Our algorithm generates a graph consisting of road intersections (nodes) and road centerlines (edges), which is a formulation that conveniently allows the subsequent use of standard resource planning and optimization techniques. Furthermore, utilizing movement data, as opposed to e.g., aerial images facilitates identification of temporary roads which are typically less visible or distinguishable in the terrain and does not require frequent collection of aerial images to capture the dynamic changes.

\section{Related work}

Automated road inference plays a pivotal role in various applications, such as urban planning, transportation management, and autonomous driving. The primary machine learning approaches for map generation are either from images, e.g., aerial photos, drone-, or satellite imagery \cite{Li2019, 8578594, Mattyus_2017_ICCV, BMVC2016_118, he2022td, SIEBERT20141}, or from movement data, such as GPS trajectories \cite{10.1145/3234692, ijgi4042446, Prabowo_2019, FU2017237, 10.1007/978-3-642-15300-6_5, he2018roadrunner, eftelioglu2022ring}. The latter approach becomes particularly advantageous when paths are not easily distinguishable in images, e.g., blend into the surrounding environment, or when the tracks change more frequently than the aerial imagery is collected. Moreover, GPS data collection is cost-effective compared to aerial or satellite imagery, especially for repeated acquisitions over time, since it can leverage existing devices such as mobile phones.

In the context of optimizing vehicle operations at construction sites, it is advantageous to model the road network as a graph. In such model, intersections are treated as key decision points, represented as nodes, while the roads themselves are depicted as the connecting edges between these nodes. Moreover, graph representation is more compact compared to alternatives such as raster images. We therefore direct our attention to works that detect road intersections and infer road centerlines, i.e., central paths along the roads. 

In CellNet \cite{10.1145/3234692}, the approach involves dividing the map into a grid and applying the mean-shifting algorithm on all GPS points within each cell, which results in identifying intersections, if they are present within the cell, or pinpointing the center of the stream of routes otherwise. To filter out the latter, it is verified that each potential intersection has at least three outgoing streams of routes by clustering the points within a predefined annulus around the candidate. The clustering approach is shown in \autoref{fig:step2} as we adapt the filtering and clustering approaches to our purpose as described in \autoref{sec:method}. In CellNet, the roads connecting the identified intersections are modeled using FastDTW, a fast and approximate implementation of dynamic time warping (DTW). 


Another approach to detect intersections involves initially identifying turning points by analyzing the difference in heading directions during individual trips, followed by clustering these turning points based on the Euclidean distance \cite{ijgi4042446}. To reduce the incidence of falsely identified intersections, only clusters with multiple points are retained. Additionally, the heading directions into and out of each potential intersection are compared with each other to remove road bends. Similarly to CellNet, the tracks connecting the intersections are aligned using dynamic time warping.


In COLTRANE (ConvolutiOnaL TRAjectory NEtwork) \cite{Prabowo_2019}, the authors propose a variation of the mean-shifting algorithm termed iterated trajectory mean shift (ITMS) to identify road centerlines. The road centerlines are treated as nodes and the candidate edges are inferred based on the distance of the GPS points from the centerlines. Finally, the edges are classified and pruned with a convolutional neural network with custom input features derived from 2D histograms of location and directional velocities.


The aforementioned methods have demonstrated strong performance in structured environments, such as cityscapes \cite{10.1145/3234692, ijgi4042446, Prabowo_2019} and airport tarmacs \cite{Prabowo_2019}. However, the movement patterns of vehicles at construction sites differ from these environments, and include reversing and maneuvering into specific positions for loading and unloading of the materials, navigating through narrow roads and rugged terrain, wide open areas, variable speeds and frequently stopping and starting. To address these unique challenges, we propose a multi-step method that first identifies the intersections in the road network and subsequently connects the roads, thereby bridging the gap in current methodologies for construction site applications.

\section{Problem definition}
The objective is to use GPS data to infer a 
graph representing the underlying road network, where each node of the graph represents a road intersection or an action point (loading or off-loading) which is described as a vector of latitude, longitude and altitude $I=(lat, lon, altitude)$. An intersection is defined as a point where at least three roads meet. The edges of the graph represent the roads and are defined as a series of consecutive, equidistant points $E=(lat, lon, altitude)$. An example of a graph is shown in \autoref{fig:results_map}.

\section{Data description} 
\label{sec:demodata}

The data used for demonstration consists of GPS updates of a fleet of dumper trucks operating at the E18 Bjørum-Skaret construction site in Norway over one day (5:00 a.m. UTC, September 13--14, 2022). The information is organized as time series and segmented into 612 distinct trips, where each trip begins when the truck is being loaded by an excavator and ends when a new one starts, therefore each trip includes phases of truck loading, driving, material delivery, and traveling empty to the next loading location. Every individual position update is defined by a timestamp, geographical coordinates ($lat, lon$) and supplementary details, such as vehicle speed and movement direction. Additionally, each trip is characterized by information identifying the machine, the driver and the task performed (e.g., "Transportation of rock from tunneling").  
The GPS data are recorded at varying time intervals determined by the truck velocity, which results in an approximately uniform spatial resolution. The statistics of the data are presented in \tabref{tab:data_statistics}.



\begin{table*}[t]
\begin{footnotesize}    
\centering
  \begin{tabular}{p{0.22\linewidth} | p{0.15\linewidth}| p{0.29\linewidth}| p{0.22\linewidth}} 

 \hline
  & Median & Mean and std. & Range \\ \hline
 Timestep & 2\,s & 4.63\,s $\pm$ 2.32\,min & [0.88\,s, 11.5\,h]\\
 Velocity (km/h) & 8.33 & 9.36 $\pm$ 6.00 & [0.0003, 25.93] \\
Distance (m) & 17.13 & 34.97 $\pm$ 277.36 &  [0, 61241]\\
Nr points/trip & 74 & 122.32 $\pm$ 188.90 & [10, 3516] \\

 \hline
 \end{tabular}
 \caption{Data statistics: Median, mean, standard deviation (std.) and range within trips.}
  \label{tab:data_statistics}
\end{footnotesize}
\end{table*}

The data is subject to noise from several sources, including the manual nature of activity logging (such as a driver pushing a button on an iPad after loading and off-loading activities), loss or weakening of the GPS signal caused by physical obstruction (e.g., passing through tunnels), inference, or bandwidth limitations. 


\section{Data preprocessing}
\label{sec:data_preprocessing}
To enhance the quality of the data, the the following data preprocessing steps are performed:
 \begin{enumerate}
 \item The Haversine formula is applied to convert latitude and longitude into coordinates expressed in meters $(x,y)$, with the origin arbitrarily set to the minimum values of latitude and longitude in the data set (\ref{app:haversine}).

  \item Incorrect or irrelevant GPS position updates are discarded, with a specific focus on the following scenarios:
  \begin{itemize}
      \item Longitude or latitude is zero: These usually occur when the GPS device is not calibrated.
      \item Velocity is zero: These are irrelevant for intersection identification and road inference as they do not represent movement. They are retained in the identification of the load and drop-off locations.
      \item Trip from the loading to drop-off point has fewer than $N_\mathrm{min}$ updates: These lack the resolution necessary for precise road inference. 
      \item The updates occur at the conclusion of trips (trip endpoints): These are often noisy because of two reasons: a) the vehicles move slowly right before loading/unloading, which creates a challenge in refining the location by the GPS receiver\footnote{GPS location is refined in the process called triangulation, which is based on the object movement.}, and b) the previous trip is considered ongoing until the truck is being loaded again, which can result in arbitrary movement around the loading site at the end of the trip. 
      \item Consecutive updates are duplicated: These do not improve the road inference but they increase the computational cost.
  \end{itemize}

 \item Each trip is interpolated with a predefined resolution $\Delta D_\mathrm{interp}$.
\end{enumerate}

\noindent The hyperparameters and their default values are given in \ref{app:hyperparameters}.

\begin{figure}[t]
  \centering
  \includegraphics[width=0.85\columnwidth]{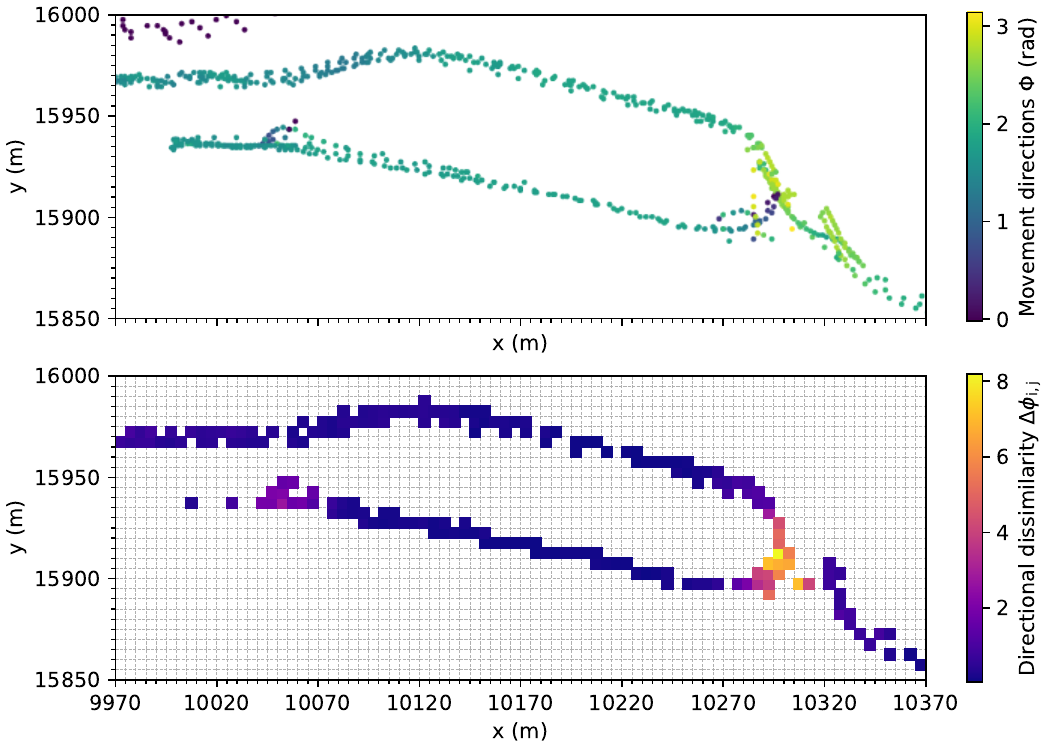}
  \caption{Identification of candidate intersections. Upper plot: The GPS trajectories are converted into movement directions $\Phi \in [0, \pi]$ (bi-directional movement, \autoref{sec:2d_hist}). Bottom plot: The area is split into grid cells, with the median direction determined for each cell. The median direction is then used to compute the directional dissimilarity $\Delta \phi _{i,j}$ between adjacent cells. The cells with $\Delta \phi _{i,j}\geq \Delta \phi_\text{thr}$ are identified as intersection candidates.}
  \label{fig:step1}
\end{figure}

\section{Graph inference algorithm}

The complete flowchart of the steps is provided in \ref{app:flowchart}.

\label{sec:method}


\subsection{2D histograms of heading directions}
\label{sec:2d_hist}

In the first step, the map is partitioned into a grid (\autoref{fig:step1}). Each cell is of size $N_\mathrm{res} \times N_\mathrm{res}=5\times 5$\,m$^2$, and is denoted as $c_{i,j}$, where $i$ and $j$ are the row and column indices, respectively. The GPS trajectories, divided into trips, can then be converted into movement directions $\Phi=\{\phi_1, \phi_2, ..., \phi_N\}$ based on the latitude and longitude of consecutive GPS updates (similar to \cite{ijgi4042446}). These directions are normalized to a $[0,\pi]$ range ($0$ to $180^{\circ}$), accounting for bi-directional vehicle movement and treating travel in either direction along the same path as equivalent, i.e.:
\begin{equation}    
\phi_{\text{norm}} = 
\begin{cases} 
\phi & \text{if } 0 \leq \phi \leq \pi, \\
2\pi - \phi & \text{if } \pi < \phi < 2\pi.
\end{cases}
\end{equation}

Finally, for each set $\Phi_{i,j} = \{\phi_{\text{norm},i,j,1}, \phi_{\text{norm},i,j,2}, \dots, \phi_{\text{norm},i,j,N}\}$ of points recorded within each cell $c_{i,j}$, the median $\tilde{\phi}_{i,j}$ is computed. This procedure essentially transforms the grid into a 2D histogram, where each cell represents the median of bi-directional vehicular movement.

\subsection{Identification of candidate intersections}
\label{sec:candidates}

Next, we identify potential intersection locations within the grid. These are hypothesized to be located in grid cells that have a notable deviation in vehicular movement direction relative to their adjacent cells, which is indicative of a directional change as vehicles pass through that cell. 

To quantitatively identify such locations, for each cell $c_{i,j}$, we compute the directional dissimilarity $\Delta\phi_\mathrm{i, j}$ defined as the root mean squared difference in the median direction between $c_{i,j}$ and its neighboring cells: 
\begin{equation}
    \Delta\phi_{i, j} = \sqrt{\sum_{k,l\in \text{neighbors}(i,j)} (\tilde{\phi}_{i,j} - \tilde{\phi}_{k,l})^2}.
\end{equation}

Here, $\text{neighbors}(i,j)$ refers to the set of indices for cells in the neighborhood of $c_{i,j}$. A cell is considered a neighbor of $c_{i,j}$ if its center lies within a distance of $D_\mathrm{nbr}=20$\,m or less from the center of $c_{i,j}$. Furthermore, only those cells that contain GPS data are taken into account. This approach yields higher values of $\Delta\phi_{i, j}$ for cells surrounded by numerous cells containing GPS points, which aligns with the expectation that intersections are more likely to appear in cells surrounded by many GPS tracks, rather than along well-defined straight roads, where multiple neighbors are likely to lack tracks.

A cell $c_{i,j}$ is flagged as a preliminary intersection candidate if its corresponding $\Delta\phi_{i, j}$ exceeds a predefined threshold $\Delta \phi_\mathrm{thr}$. With the default cell resolution of $N_\mathrm{res}=5$\,m, this typically results in multiple adjacent cells being flagged as potential intersections. To refine these results, the neighboring candidates are further merged with a simple clustering approach. Candidates within a distance of $D_\mathrm{int\_clust}=15$\,m from each other are grouped together, and the central point of each group is marked as the final intersection candidate. Preliminary candidates that do not form a part of any cluster are excluded under the assumption that the chosen $N_\mathrm{res}$ is small enough for the intersections and their immediate surroundings to extend over several cells. 




\begin{figure}[t!]
  \centering
  \includegraphics[width=0.85\columnwidth]{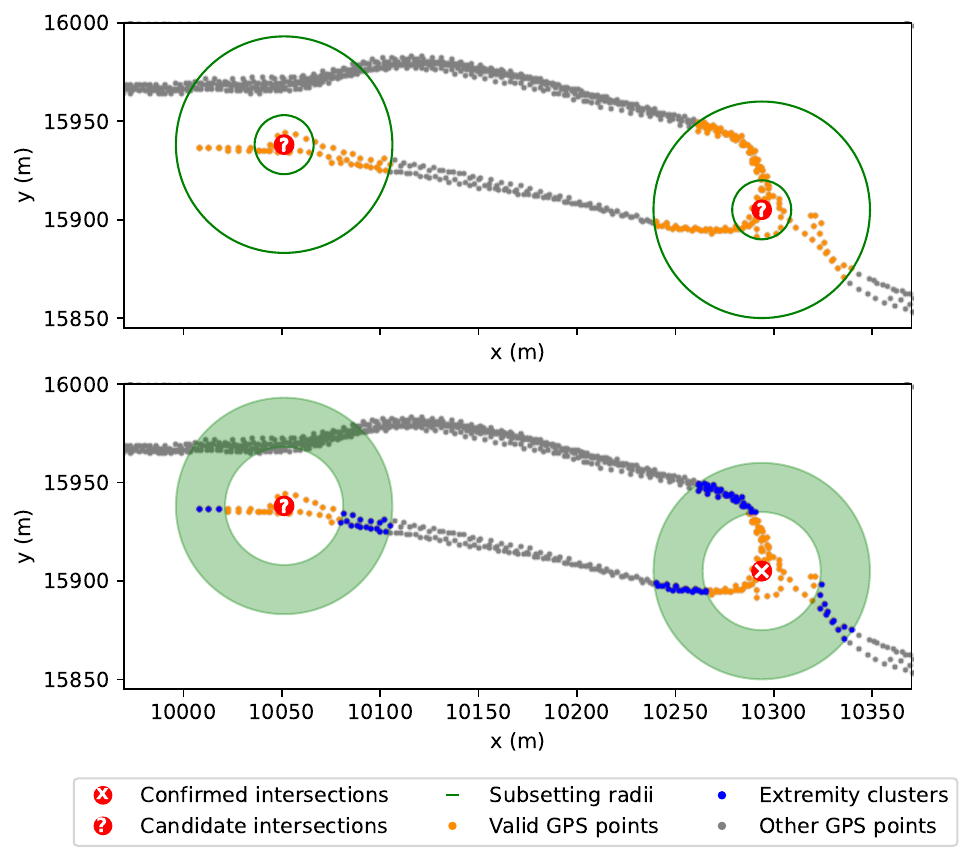}
  \caption{Validation of intersection candidates. Upper plot: the points located within the outer radius $R+L$ from each intersection candidate are clustered using DBSCAN. Those points that belong to a cluster where at least one member is within the radius $D_\mathrm{passing}$ from an intersection candidate are considered "valid points" for the next step. Bottom plot: The "valid points" are subset by annuli with radii $R$ and $R+L$ and clustered with DBSCAN. The left candidate is rejected, as only two clusters leaving the intersection are detected, while the right candidate is accepted as it has three clusters leaving the intersection. Note that the method is robust against the presence of parallel roads that are not connected to the intersection.}
   \label{fig:step2}
\end{figure}

\subsection{Validation of intersections}
\label{sec:inter_validation}

The next step consists of filtering out false positives among the candidate intersections using an adapted version of the method proposed in \cite{10.1145/3234692}. This method discerns intersections from e.g., road bends by determining the number of roads leaving the candidate location for the intersection.
In our version, we first subset the GPS points to those that lie within the circle bounded by the outer radius $R+L$ and apply DBSCAN clustering to those points, with the clustering radius of $\epsilon_\mathrm{passing}=12$\,m and the minimum sample size $N_\mathrm{passing}=5$. The points that are not directly connected to the intersection, i.e., they do not form a cluster with the points passing next to the candidate intersection within a distance of $D_\mathrm{passing}=15$\,m, are filtered out. Next, following \cite{10.1145/3234692}, we define an extremity annulus bounded by two concentric circles with radii $R$ and $R+L$ around each candidate intersection (\autoref{fig:step2}).
The GPS points located within the annulus and are connected to the intersection are clustered with a clustering distance of $D_\mathrm{ext\_clust}=20$\,m. Each cluster of a size larger than $N_\mathrm{ext\_clust}=5$ represents a road that passes through the extremity. The final set of intersections are those that have at least three such clusters in their extremity, representing three roads leaving the intersection. The values of $R$ and $L$ can vary depending on the actual size of intersections, and this step can be repeated with multiple values, e.g., $R=[30,100]$ and $L=R+25$.

 To reduce computations for determining distances between the GPS points and intersections, we employed k-dimensional trees (k-d trees) that partition space into a hierarchical tree structure, significantly accelerating spatial queries. This approach allowed us to narrow down calculations to only those pairs of GPS points and intersection centers that are within a predefined maximum distance $R_\mathrm{max}+L$ from each other. Moreover, while our application did not show significant differences across clustering techniques, HDBSCAN may be a suitable alternative in cases where handling varying densities of GPS data is necessary.

\begin{figure}[t!]
  \centering
  \includegraphics[width=0.85\columnwidth]{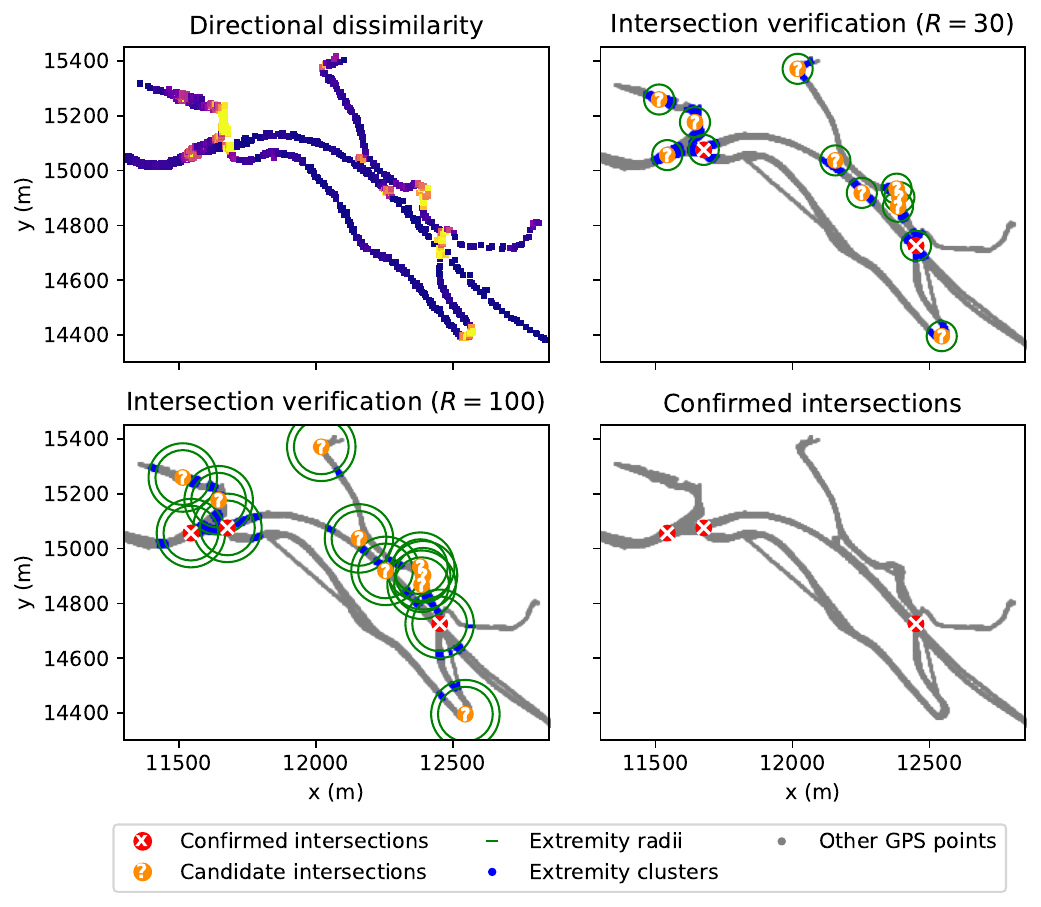}
  \caption{Step-by-step intersections detection and validation in Zone 1. Upper left: The flare-ups in directional dissimilarity suggest possible intersections. Upper right: The intersection candidates are verified with an annulus of a small radius $R=30$. Road bends and vehicle turning points that do not have three outward roads are discarded. Bottom left: The step is repeated with a larger radius $R=100$, allowing to find the third intersection. Bottom right: Only three intersections are confirmed, which is the correct result.}
   \label{fig:details_detection}
\end{figure}

\subsection{Identifying load and drop-off locations}
\label{sec:load_dropoff}
Load and drop-off locations are also considered nodes. Although load and drop-off locations are reported in the data for each individual trip, due to the manual nature of the logging process, they are prone to uncertainty. To minimize the uncertainty, the reported locations are clustered as follows:
\begin{itemize}
   \item Loading points: The reported loading points are organized by excavator ID and task ID. Within each group, the loading point is determined as the trimmed average of all positions. The loading points that are closer to each other than a predefined threshold, $D_\mathrm{load/dump}=100$\,m, are merged.
   \item Drop-off points: The reported drop-off points are clustered using density-based clustering (DBSCAN\,\cite{dbscan}) 
    and the mean $(lat,lon)$ within each cluster is used to compute the drop-off location. The locations that are closer than $D_\mathrm{load/dump}=100$\,m are merged.
\end{itemize}


\subsection{Road inference}
\label{sec:road_inference}

In the final step of constructing the road network, the nodes, consisting of intersections, loading and drop-off locations, are interconnected by edges that represent the roads. Additional edges are also created to depict roads with a dead end, i.e., roads that pass through only one intersection. 

The process starts by identifying the sections of the trip that pass in proximity to any node, i.e., within a predefined radius of $D_\mathrm{node}=30$\,m. For each of these sections, we mark the time step at which the GPS points are the closest to that node. The trips are then split into smaller segments at these time steps, resulting in segments that start at either the beginning of the trip or at a node, and similarly finish either at the end of the trip or at a node. The resulting segments are further grouped by their adjacent node pairs or singular nodes. 

Following this, the segments in each group are assigned to clusters. The number of clusters determines the number of roads that are present in each group. The process is as follows: Firstly, all segments are trimmed around the nodes by $D_\mathrm{node}=30$\,m (or optionally a larger distance) to avoid the separate roads connecting at these points. The GPS points within each group are clustered by their $(x,y)$ coordinates using DBSCAN. 
Subsequently, each segment is assigned its dominant cluster, defined as the cluster that appears the most frequently within this segment. For each dominant cluster, a representative segment is selected as an edge. This selection is made at random among the segments that have the median number of discretization points within the specified cluster. The final graphs consist of a list of nodes and edges that either interconnect two nodes or are connected to a node.

\begin{figure*}[t!]
\begin{center}
\begin{subfigure}{0.45\textwidth}
\includegraphics[width=\columnwidth]{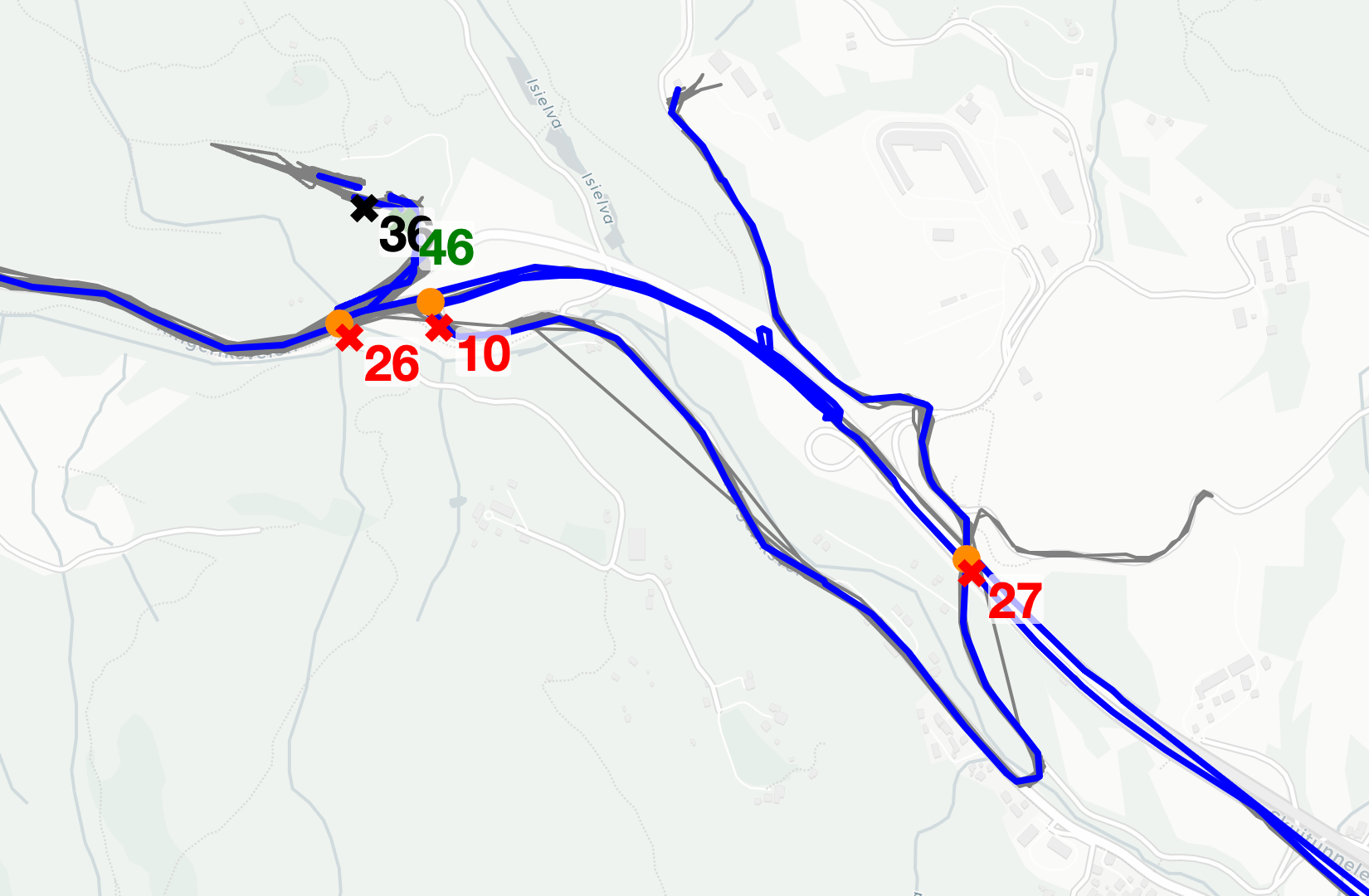}
\caption{Zone 1.}
\label{fig:results_map_a}
\end{subfigure}
\begin{subfigure}{0.45\textwidth}
\includegraphics[width=\columnwidth]{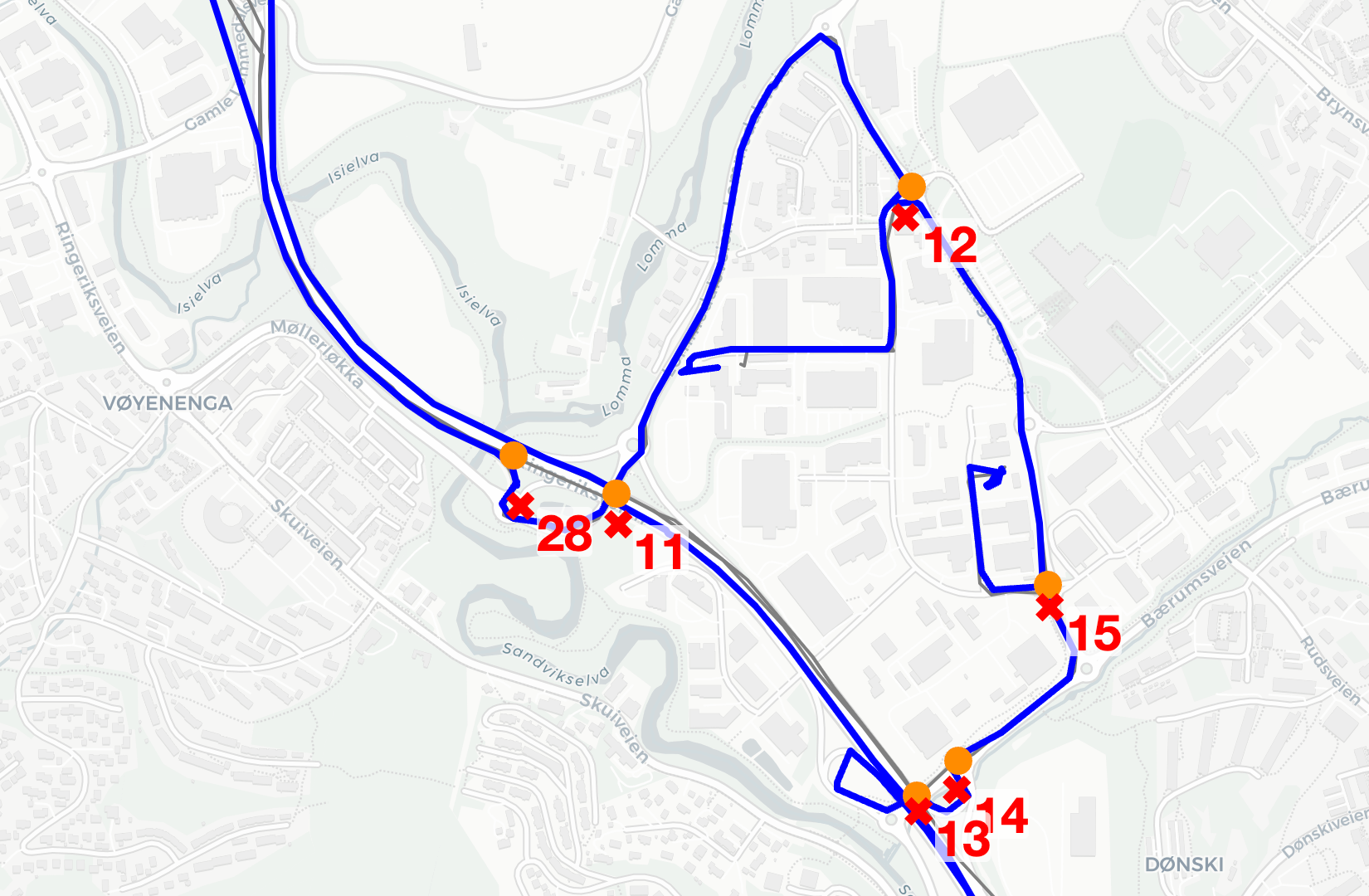}
\caption{Zone 2.}
\label{fig:results_map_b}
\end{subfigure}
\begin{subfigure}{0.45\textwidth}
\vskip 0.2in
\includegraphics[width=\columnwidth]{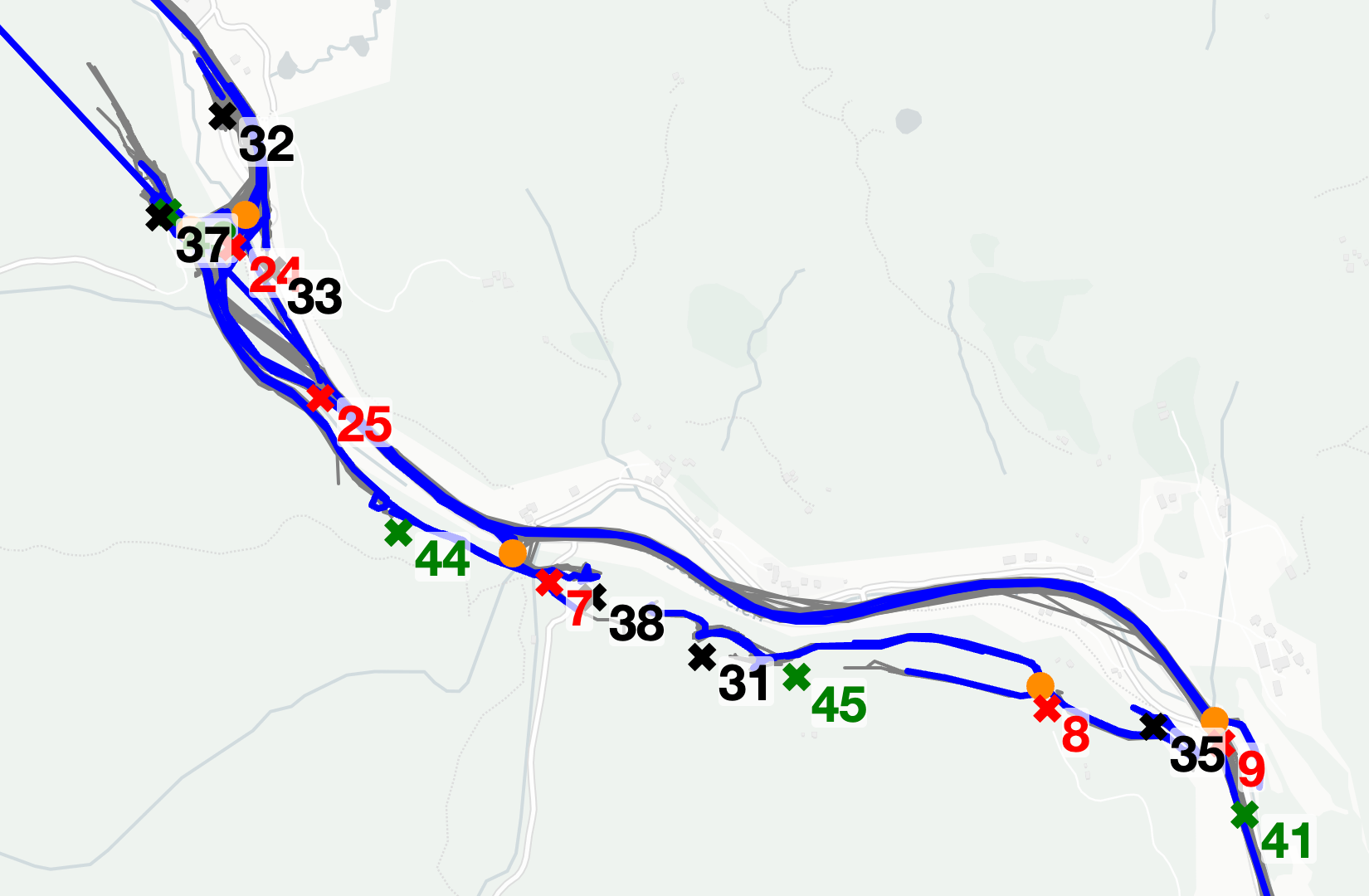}
\caption{Zone 3.}
\label{fig:results_map_c}
\end{subfigure}
\begin{subfigure}{0.45\textwidth}
\vskip 0.2in
\includegraphics[width=\columnwidth]{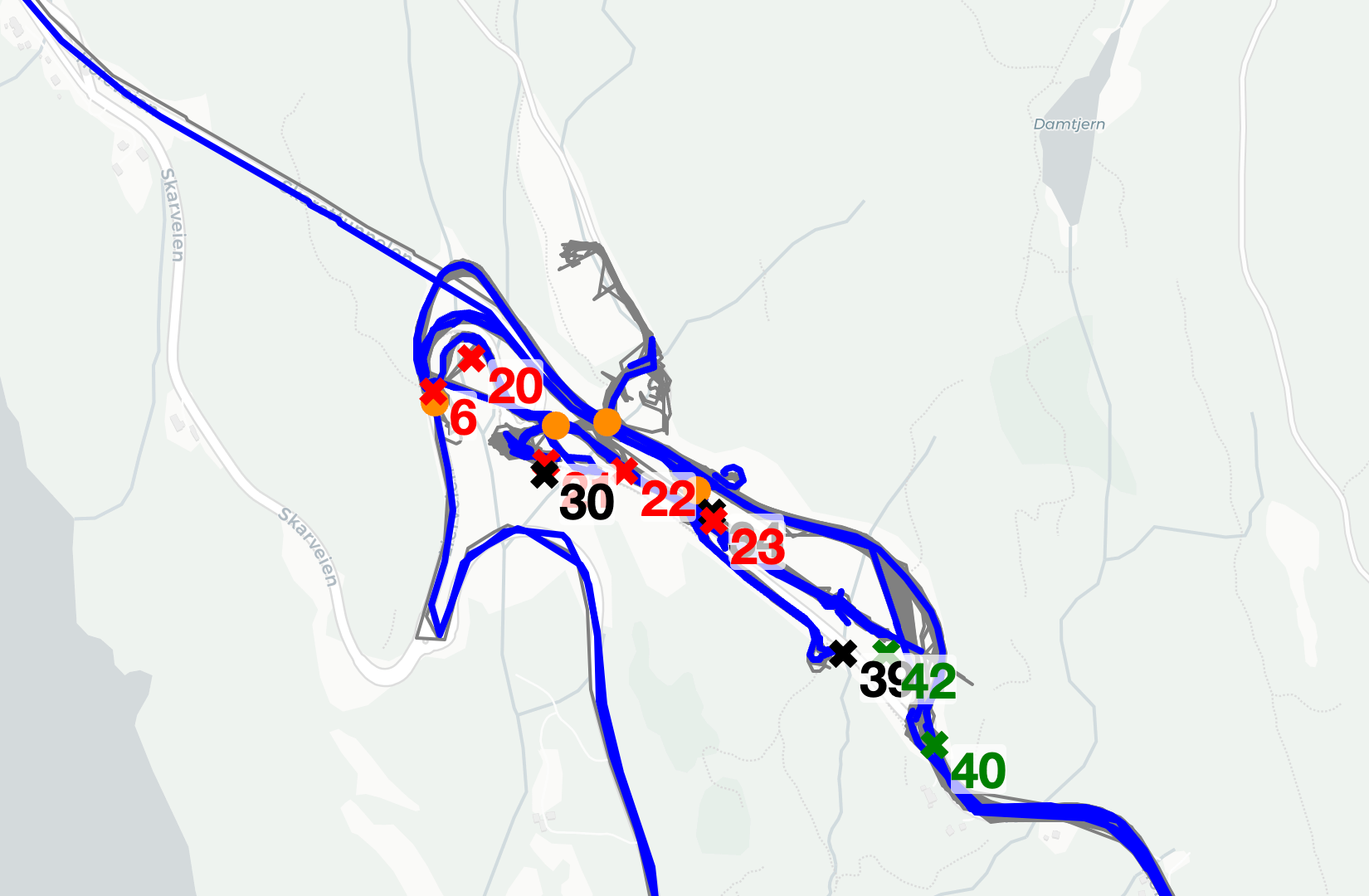}
\caption{Zone 4.}
\label{fig:results_map_d}
\end{subfigure}
\caption{Examples of the inferred graph with crosses representing three types of nodes: intersections (red), load points (green), and drop-off points (black). The actual intersections are represented with orange points. The edges of the graph, i.e., the roads are in blue, while the actual GPS data are represented by gray lines.
}
\label{fig:results_map}
\end{center}
\vskip -0.2in
\end{figure*}

\begin{figure*}[t!]
  \centering
  \includegraphics[width=0.95\columnwidth]{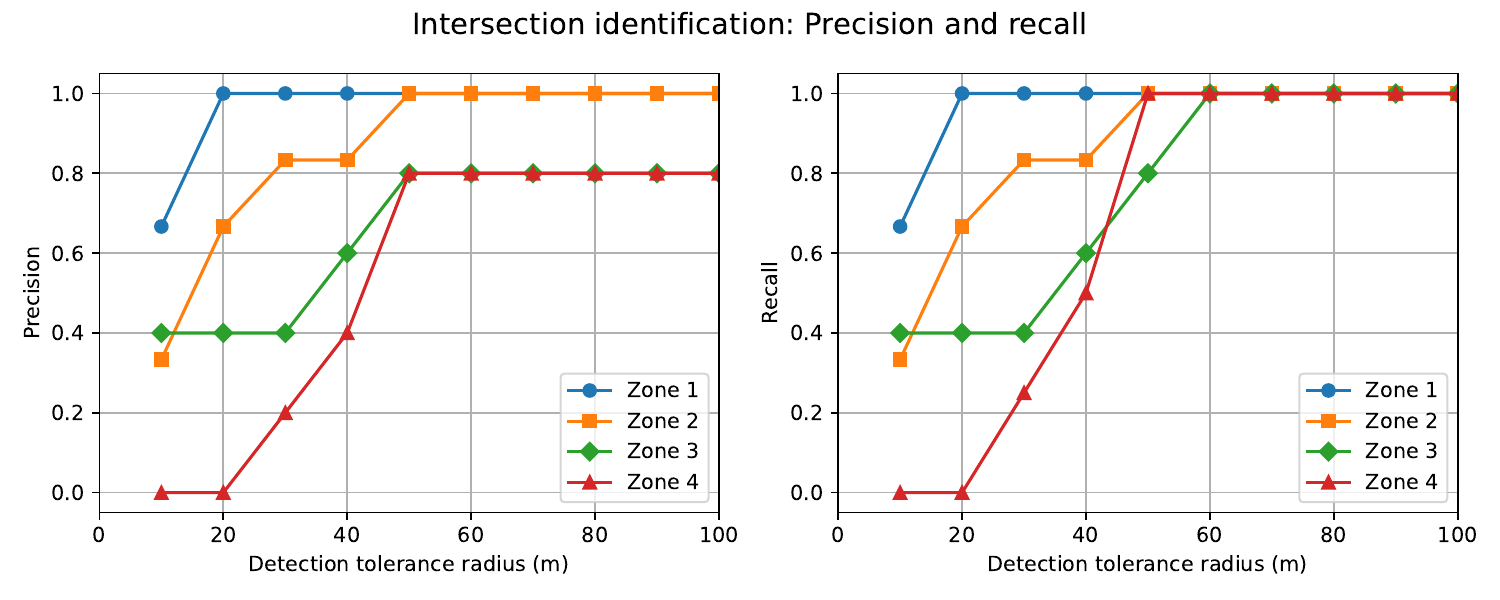}
  \caption{Precision and recall for detecting intersections in the considered zones at varying distance thresholds. The \textit{detection tolerance radius} is the radius around a labeled intersection within which a predicted intersection is considered detected.}
   \label{fig:pr_dist}
\end{figure*}

\section{Results}

The visual representation of the performance of the algorithm is presented in \autoref{fig:results_map}. Each of the four subplots shows segments of the inferred graph in the order from relatively straightforward to more complex scenarios. The parameters used are defined in \ref{app:hyperparameters}. The intersection validation step was run twice, with $R=30$\,m and $R=100$\,m (\autoref{fig:details_detection}).

In the first two examples: Zone 1 and 2, the algorithm achieves perfect recall and precision: All intersections are detected correctly and all driven roads are outlined. The use of at least two radii in the intersection validation step is crucial to detect cross-roads of varying shape and spread, e.g., merging roads (26, Zone 1) and small detours (14, Zone 2). We note that the algorithm is robust against noisy GPS updates present in Zone 1 (visible as straight lines). This robustness against noise that visually resembles a road split is gained when quantifying the directional dissimilarity (\autoref{sec:candidates}), as the area is never qualified as an intersection candidate (see upper left in \autoref{fig:details_detection}). 

In our further analysis, we show the limitation of the algorithm in the presence of consistent noise. In Zone 3 four intersections are detected correctly (7, 8, 9, and 24). Number 25 is falsely identified as an intersection due to the noisy data. In this case, noisy updates, visible as straight gray lines, are more frequent than the correct outline of the road. Zone 4 presents the biggest challenge for the algorithm, with noisy data dominating several parts of the map, in the areas where vehicles often change direction. In this case, the algorithm correctly detects intersection 6, identifies some intersections near the actual ones (e.g., 21 and 23), and produces several false positives (e.g., 20).  

The performance is further quantified with precision and recall (definition in \ref{app:metrics}), as shown in \autoref{fig:pr_dist}. Both metrics improve as the distance threshold increases, reflecting a more relaxed detection criterion. However, the level of alignment between the actual and predicted intersections varies between zones. In Zone 1, high precision and recall (\(75\%-100\%\)) are achieved even at a threshold of 10–20\,m, indicating tight detection. In contrast, Zone 4 requires a threshold of up to 50\,m to reach similar values, suggesting a lower level of alignment. Zones 2 and 3 exhibit intermediate behavior. In Zone 2, precision and recall reach $90\%$ at 30\,m and $100\%$ at 50\,m. For Zone 3, both recall and precision start at $50\%$, eventually reaching $100\%$ recall and $83\%$ precision at at 50\,m.

\section{Discussion}

While the proposed method was demonstrated to efficiently map the structure of large road networks and finding intersections, the accuracy of the method depends on two factors:
\begin{itemize}
    \item[a)] Contextual parameter tuning: The algorithm's parameters have to be tuned for the specific application and characteristics of the landscape. For instance, the distance parameters that define the relevant area surrounding a singular point marked as an intersection will vary depending on the area covered by an intersection: a well-defined crossroad will need tighter parameter settings than more expansive roundabouts or open-area nodes with several diverging routes.r
    \item[b)] Quality of the GPS data: While multiple problems with the data can be addressed with the steps outlined in \autoref{sec:data_preprocessing}, mitigating certain types of noise is not trivial, especially if the inaccurate GPS updates are dominant and overshadow valid data. In such cases, customizing the data cleansing process to the specific deployment context becomes crucial. Additionally, employing techniques to enhance GPS position accuracy can further improve data quality and mitigate the impact of noise \cite{aggarwal2020enhancement, kumar2014improving, kumar2013global, kumar2014identifying}.
\end{itemize}

\section{Conclusions}

We have introduced a novel method for road inference from GPS data, which constructs a detailed graph of road networks with nodes representing intersections, load points, and drop-off points, while edges correspond to the connecting roads. The algorithm demonstrates a high level of proficiency in mapping large-scale road networks and identifying road intersections and achieves perfect precision and recall when data quality is high. However, it can be sensitive to noise, particularly for irregular GPS updates, e.g., when the signal is lost as a vehicle passes through a tunnel. Our approach incorporates a selection of physically meaningful and understandable parameters that offer flexibility for tuning across diverse terrains. This automated graph construction approach will allow for improved coordination and management of units at construction sites.

\section*{Acknowledgements}

This work is based upon the support from the Research Council of Norway under the Datadrevet Anleggsplass 309797 (English: Data-driven Construction Site). Data were collected by Skanska Norway AS and accessed through the \href{https://ditioapp.com/}{Ditio app (https://ditioapp.com/)}. 

\section*{Author contributions}

\begin{itemize}
    \item Katarzyna Michałowska: Conceptualization, methodology, data curation, software, investigation, formal analysis, validation, visualization,  writing (original draft),
    \item Helga Margrete Bodahl Holmestad: Conceptualization, methodology, data curation, software, investigation, visualization, writing (review and editing),
    \item Signe Riemer-Sørensen: Conceptualization, methodology, writing (review and editing), resources, project administration, funding acquisition.
\end{itemize}

\section*{Data and code availability}

The data used in this study are proprietary to Skanska AS and can be made available upon reasonable request. The code developed in this study is available in this \href{https://github.com/katarzynamichalowska/road-graph-generator}{GitHub repository: https://github.com/katarzynamichalowska/road-graph-generator}.

\section*{Conflict of interest}
The authors declare no conflict of interest. 

\section*{Declaration of generative AI and AI-assisted technologies in the writing process}

During the preparation of this work the authors used \href{https://chat.openai.com/}{ChatGPT} to improve the clarity of the language in certain sections of the manuscript. After using this tool, the authors reviewed and edited the content as needed and take full responsibility for the content of the publication.





\printbibliography

\newpage
\FloatBarrier
\appendix
\section{Nomenclature}

\begin{table*}[h!]
\begin{footnotesize}
    \centering
  \begin{tabular}{p{0.13\linewidth} | p{0.82\linewidth}} 
 \hline
 \textbf{Notion} & \textbf{Explanation}  \\ \hline
 GPS & Global positioning system. \\
 Intersection & A geographical point where at least three roads meet.\\
 Trip & A sequence of consecutive GPS data points representing the movement of a truck from its loading point until it is reloaded, signifying a complete transportation cycle. \\ 
$(lat, lon)$ & Coordinates expressed as latitude and longitude. \\
 $(x,y)$ & Coordinates expressed in meters with the origin set to the minimal values of latitude and longitude in the dataset. \\
 DTW & Dynamic time warping.\\
 \hline
 \end{tabular}
 \caption{Nomenclature used in the paper.}
  \label{tab:nomenclature}
\end{footnotesize}    
\end{table*}

\section{Metrics}
\label{app:metrics}
\subsection{Precision and recall}
\begin{equation}
\text{Precision} = \frac{\text{TP}}{\text{TP} + \text{FP}}
\end{equation}

\begin{equation}
\text{Recall} = \frac{\text{TP}}{\text{TP} + \text{FN}}
\end{equation}

\noindent where:
\begin{itemize}
    \item \textbf{TP (True Positives)}: Predicted intersections that correctly match actual intersections within a defined distance threshold.
    \item \textbf{FP (False Positives)}: Predicted intersections that do not correspond to any actual intersection.
    \item \textbf{FN (False Negatives)}: Actual intersections that were not detected by the algorithm.
\end{itemize}

\section{Mathematical formulas}
\label{app:haversine}
\subsection{Haversine formula}

The Haversine formula, used for calculating the distance between two points on the surface of a sphere given their latitude and longitude, is given by the following:
\[
a = \sin^2\left(\frac{\Delta\phi}{2}\right) + \cos(\phi_1) \cos(\phi_2) \sin^2\left(\frac{\Delta\lambda}{2}\right)
\]
\[
c = 2 \arctan2\left(\sqrt{a}, \sqrt{1-a}\right)
\]
\[
d = R \cdot c
\]

\noindent where:
\begin{itemize}
    \item $\Delta\phi  = \phi_2 - \phi_1 \quad (\text{latitude difference in radians})$
    \item $\Delta\lambda  = \lambda_2 - \lambda_1 \quad (\text{longitude difference in radians}) $
    \item $R  = \text{Earth's radius (6,371 km)}$
    \item $d  = \text{Distance between the two points in km (along the surface of the sphere)}$
\end{itemize}

\clearpage
\section{Method flowchart}
\label{app:flowchart}

\begin{figure}[h!]
\centering
\begin{tikzpicture}[
    node distance=1cm and 0.8cm,
    box/.style={rectangle, draw, rounded corners, text width=4.5cm, align=left, minimum height=4cm, font=\footnotesize},
    sectionnum/.style={font=\bfseries\small, rectangle, fill=white, inner sep=2pt, text=black}
]

\node[box, minimum height=4.8cm, align=left] (data) {
    \textbf{\small Preprocessing}
    \begin{compactitem}
        \item Convert coordinates into metres.
        \item Remove noisy GPS updates.
        \item Interpolate paths.
    \end{compactitem}
    \textit{Parameters:} $D_\text{endpoints}, \Delta D_\text{interp}, N_\text{min}$
};
\node[sectionnum] at ($(data.north) + (0, 0.5cm)$) {Section \ref{sec:data_preprocessing}};

\node[box, right=of data, minimum height=4.8cm, align=left] (histogram) {
    \textbf{\small 2D histograms} \\
    \begin{compactitem}
        \item Partition the area into a grid.
        \item Compute the median heading directions in each cell based on the GPS updates.
    \end{compactitem}
    \textit{Parameters:} $N_\text{res}$
};
\node[sectionnum] at ($(histogram.north) + (0, 0.5cm)$) {Section \ref{sec:2d_hist}};

\node[box, right=of histogram, minimum height=4.8cm, align=left] (intersections) {
    \textbf{\small Intersection identification} \\
    \begin{compactitem}
        \item Compute the directional dissimilarity in each cell.
        \item Flag intersection candidates with the largest dissimilarity.
    \end{compactitem}
    \textit{Parameters:} $D_\text{nbr}, D_\text{int\_clust}, \Delta\phi_\text{thr}$
};
\node[sectionnum] at ($(intersections.north) + (0, 0.5cm)$) {Section \ref{sec:candidates}};

\node[box, below=of data, yshift=-1cm, minimum height=4.8cm] (validate) {
   \textbf{\small Intersection validation} \\
   \begin{compactitem}
       \item Cluster points in extremities.
       \item Validate outgoing roads using those clusters.
   \end{compactitem}
    \textit{Parameters:} $R, L,$ $\epsilon_\mathrm{passing}$, $N_\mathrm{min\_passing}$, $D_\text{passing}, D_\text{ext\_clust}, N_\text{ext\_clust}$
};
\node[sectionnum] at ($(validate.north) + (0, 0.5cm)$) {Section \ref{sec:inter_validation}};

\node[box, right=of validate, minimum height=4.8cm] (loading) {
    \textbf{\small Load/drop-off detection} \\
    \begin{compactitem}
        \item Cluster load and drop-off locations.
        \item Merge points that are close together.
    \end{compactitem}
    \textit{Parameters:} $D_\text{load/dump}$
};
\node[sectionnum] at ($(loading.north) + (0, 0.5cm)$) {Section \ref{sec:load_dropoff}};

\node[box, right=of loading, minimum height=4.8cm] (inference) {
    \textbf{\small Road inference} \\
    \begin{compactitem}
        \item Connect the nodes with GPS paths.
        \item Cluster GPS segments.
        \item Assign edges to the nodes.
    \end{compactitem}
    \textit{Parameters:} $D_\text{node}, \epsilon_\text{road}, N_\text{min\_road}$
};
\node[sectionnum] at ($(inference.north) + (0, 0.5cm)$) {Section \ref{sec:road_inference}};

\node[box, below=of inference, xshift=-5cm, minimum height=3cm] (graph) {
    \textbf{\small Final output} \\
    \begin{compactitem}
        \item \textbf{Nodes}: Intersections, load/drop-off points
        \item \textbf{Edges}: Roads connecting nodes
    \end{compactitem}    
};

\draw[-{Latex}, thick] (data) -- (histogram);
\draw[-{Latex}, thick] (histogram) -- (intersections);
\draw[-{Latex}, thick] (intersections.south) -- ++(0, -0.8) -| (validate.north);
\draw[-{Latex}, thick] (validate) -- (loading);
\draw[-{Latex}, thick] (loading) -- (inference);
\draw[-{Latex}, thick] (inference.south) -- ++(0, -0.8) -| (graph.north);

\end{tikzpicture}
\caption{The flowchart of the full road graph generation framework.}
\label{fig:framework_sections}
\end{figure}

\FloatBarrier
\clearpage
\section{Hyperparameters for algorithm and data processing}
\label{app:hyperparameters}

\begin{table*}[h!]
\begin{footnotesize}    
\begin{tabular}{p{0.15\linewidth} | p{0.8\linewidth}} 
\multicolumn{2}{l}{\textbf{Data preprocessing parameters}} \\ \hline
\textbf{Parameter} & \textbf{Description and tuning guidance} \\ \hline

\textbf{$D_\mathrm{endpoints}$} & 
Distance driven removed at the end of each trip to reduce noise. 
\newline \textbf{Default}: 100\,m. 
\newline \textbf{Tuning}: Increase to discard noisy points at trip endpoints, especially if GPS accuracy decreases at low speed. Decrease to retain more data near loading/unloading points. \\ \hline

\textbf{$d_\mathrm{spline}$} & 
Spline degree used for interpolation. Higher degrees may lead to overfitting and inaccuracies.
\newline \textbf{Default}: 1 (linear interpolation). 
\newline \textbf{Tuning}: Increase for smoother paths, but be cautious of overfitting to noise. Use 1 for reliable, simple interpolation. \\ \hline

\textbf{$\Delta D_\mathrm{interp}$} & 
Resolution for interpolating GPS paths.
\newline \textbf{Default}: 5\,m. 
\newline \textbf{Tuning}: Increase to reduce the computational cost and memory. Keep or decrease to retain detail in the paths. \\ \hline

\textbf{$N_\mathrm{min}$} & 
Minimum number of data points required to define a valid trip.
\newline \textbf{Default}: 10 points. 
\newline \textbf{Tuning}: Increase to exclude very short trips irrelevant for road inference. Decrease to capture brief trips if necessary. \\ \hline

\textbf{$\Delta t_\mathrm{gap}$} & Time threshold to split trips if the vehicle is idle for too long.
\newline \textbf{Default}: 5\,min. 
\newline \textbf{Tuning}: Increase if frequent long pauses should count as part of the trip. Decrease to reflect shorter idles as separate trips. \\ \hline

\textbf{$\Delta d_\mathrm{gap}$} & Distance threshold to split trips if consecutive GPS updates are too far apart, indicating a noisy signal. 
\newline \textbf{Default}: 2000\,m. 
\newline \textbf{Tuning}: Increase for large distances with sparse updates that need connecting. Decrease for noisy updates causing errors in road segments.
 \\ \hline

\textbf{$\Delta \phi_\mathrm{gap}$} & Angular threshold to split trips based on directional changes.
\newline \textbf{Default}: 0$^{\circ}$. 
\newline \textbf{Tuning}: Increase if significant directional changes accurately represent the road layout. Decrease to split trips at sharp turns (e.g., U-turns) for better road inference.\\ \hline

\end{tabular}
\caption{Parameters used in data preprocessing.}
\label{tab:params1}
\end{footnotesize}
\end{table*}

\begin{table*}[h!]
\begin{footnotesize}    
\begin{tabular}{p{0.15\linewidth} | p{0.8\linewidth}} 
\multicolumn{2}{l}{\textbf{2D histograms of heading directions}} \\ \hline
\textbf{Parameter} & \textbf{Description and tuning guidance} \\ \hline
\textbf{$N_\mathrm{res}$} & 
Grid cell resolution for the 2D histogram construction of heading directions. \newline
\textbf{Default}: 5\,m. 
\newline \textbf{Tuning}: Increase to speed up processing by reducing the cell count. Decrease for finer granularity in detecting directional changes.\\ \hline
\end{tabular}
\caption{Parameters used in 2D histograms of heading directions.}
\label{tab:params0}
\end{footnotesize}
\end{table*}

\begin{table*}[h!]
\begin{footnotesize}    
\begin{tabular}{p{0.15\linewidth} | p{0.8\linewidth}} 
\multicolumn{2}{l}{\textbf{Identification of candidate intersections}} \\ \hline
\textbf{Parameter} & \textbf{Description and tuning guidance} \\ \hline
\textbf{$D_\mathrm{nbr}$} & 
Maximum distance between grid cell centers to consider them as neighbors. Neighboring cells are used to assess directional dissimilarity between the intersection center and its neighbors based on the principal heading directions. \newline
\textbf{Default}: 20\,m. 
\newline \textbf{Tuning}: Increase to consider larger intersections (and vice versa).\\ \hline
\textbf{$\Delta \phi_\mathrm{thr}$} & 
Threshold for the angular difference in principal heading direction between neighboring cells to identify them as intersection candidates. \newline
\textbf{Default}: 1.4$^{\circ}$. 
\newline \textbf{Tuning}: Decrease to detect smaller angular changes indicating intersections. Increase to focus on significant direction changes. \\ \hline
\textbf{$D_\mathrm{int\_clust}$} & 
Clustering distance for merging nearby intersection candidates. \newline
\textbf{Default}: 15\,m. 
\newline \textbf{Tuning}: Increase to merge more distant intersection candidates. Decrease to separate closely located ones.\\ \hline
\end{tabular}

 \caption{Parameters used for identification of intersection candidates.}
  \label{tab:params2}
\end{footnotesize}
\end{table*}

\begin{table*}[h!]
\begin{footnotesize}    
\begin{tabular}{p{0.15\linewidth} | p{0.8\linewidth}} 
\multicolumn{2}{l}{\textbf{Validation of intersections}} \\ \hline

\textbf{Parameter} & \textbf{Description and tuning guidance} \\ \hline

\textbf{$R$} & 
Inner radius of the extremity annulus for validating intersections. \newline
\textbf{Default}: 30\,m, 100\,m. 
\newline \textbf{Tuning}: Increase to capture intersections with wider centers. Decrease to detect more compact intersections and account for smaller roads leaving the intersection. \\ \hline

\textbf{$L$} & 
Width of the extremity annulus used in clustering, with the outer radius defined as $R+L$. \newline
\textbf{Default}: 25\,m. 
\newline \textbf{Tuning}: Increase to capture more GPS points around the intersection. Decrease to avoid merging roads that connect further away into a single cluster. \\ \hline

\textbf{$N_\mathrm{max\_val}$} & Maximum number of GPS points in the extremity annulus used for each intersection validation.\newline
\textbf{Default}: 1000. 
\newline \textbf{Tuning}: Increase for potential increase in accuracy. Decrease for faster performance. \\ \hline
\textbf{$\epsilon_\mathrm{passing}$} & 
Radius for DBSCAN clustering when validating points within the R+L circles as roads passing through the intersection. \newline
\textbf{Default}: 12\,m. 
\newline \textbf{Tuning}: Increase to merge more spread-out GPS points. Decrease to avoid merging neighboring but separate roads. \\ \hline
\textbf{$N_\mathrm{min\_passing}$} & 
Minimum sample count for DBSCAN to form a road passing by the intersection (applied to the points within the R+L circles). \newline
\textbf{Default}: 5 points. 
\newline \textbf{Tuning}: Increase to require more GPS points for defining a road, which can eliminate noise but may miss roads with sparse data. Decrease to detect more roads, even those with less data coverage. \\ \hline
\textbf{$D_\mathrm{passing}$} & 
Maximum distance between at least one of the points in each cluster within R+L and the intersection center to include this cluster of points in intersection validation. \newline
\textbf{Default}: 15\,m. 
\newline \textbf{Tuning}: Increase for wide intersections to capture all trips passing through. Decrease to exclude trips passing nearby but not through the intersection. \\ \hline
\textbf{$D_\mathrm{ext\_clust}$} & 
Distance for clustering GPS points within the extremity annulus to represent outgoing roads. \newline
\textbf{Default}: 20\,m. 
\newline \textbf{Tuning}: Increase to merge distant or scattered GPS points as a single road. Decrease to prevent merging multiple roads into one. \\ \hline
\textbf{$N_\mathrm{ext\_clust}$} & 
Minimum number of points required in a cluster within the extremity annulus to consider it as a road passing through the intersection. \newline
\textbf{Default}: 5 points. 
\newline \textbf{Tuning}: Increase for stricter road validation, enhancing the reliability of the algorithm. Decrease to recognize clusters with fewer points and include less frequently traveled roads. \\ \hline
\end{tabular}
  \caption{Parameters used for validation of intersections.}
  \label{tab:params3}
\end{footnotesize}
\end{table*}

\begin{table*}[h!]
\begin{footnotesize}
\begin{tabular}{p{0.15\linewidth} | p{0.8\linewidth}} 
\multicolumn{2}{l}{\textbf{Loading and drop-off nodes}} \\ \hline
\textbf{Parameter} & \textbf{Description and tuning guidance} \\ \hline
\textbf{$D_\mathrm{load/dump}$} & 
Distance threshold for merging loading or drop-off locations. \newline
\textbf{Default}: 100\,m. 
\newline \textbf{Tuning}: Increase to combine more distant locations, resulting in larger aggregated zones. Decrease to identify more separated locations. \\ \hline
\end{tabular}
  \caption{Parameters used for identification of loading and drop-off nodes.}
  \label{tab:params4}
\end{footnotesize}
\end{table*}

\begin{table*}[h!]
\begin{footnotesize}
\begin{tabular}{p{0.15\linewidth} | p{0.8\linewidth}} 
\multicolumn{2}{l}{\textbf{Road inference}} \\ \hline
\textbf{Parameter} & \textbf{Description and tuning guidance} \\ \hline
\textbf{$D_\mathrm{node}$} & 
Maximum distance between a GPS track and a node center for the trip to be considered as passing through that node (intersection or a loading/drop-off point). \newline
\textbf{Default}: 30\,m. 
\newline \textbf{Tuning}: Increase to include roads passing further from the node center, such as at wide intersections with dispersed tracks. Decrease to exclude roads that are only nearby. \\ \hline
\textbf{$\epsilon_\mathrm{road}$} & 
Radius for DBSCAN clustering during road segment detection. \newline
\textbf{Default}: 15\,m. 
\newline \textbf{Tuning}: Increase to merge more spread-out GPS points, possibly connecting separate road segments. Decrease for tighter clustering, focusing on clearer road definitions. \\ \hline
\textbf{$N_\mathrm{min\_road}$} & 
Minimum sample count for DBSCAN to form a road segment cluster. \newline
\textbf{Default}: 5 points. 
\newline \textbf{Tuning}: Increase to require more GPS points for defining a road, which can eliminate noise but may miss roads with sparse data. Decrease to detect more roads, even those with less data coverage. \\ \hline
\end{tabular}
  \caption{Parameters used for road inference.}
  \label{tab:params5}
\end{footnotesize}
\end{table*}

\end{document}